\documentclass[conference]{IEEEtran}

\usepackage[utf8]{inputenc} 
\usepackage[T1]{fontenc}    
\usepackage{hyperref}       
\usepackage{url}            
\usepackage{booktabs}       
\usepackage{amsfonts}       
\usepackage{nicefrac}       
\usepackage{microtype}      
\usepackage{graphicx}
\usepackage{bm} 
\usepackage{amsmath}
\usepackage[ruled,vlined]{algorithm2e}
\usepackage{caption}
\usepackage{subcaption}
\usepackage{cite}

\begin{document}
\title{One Line To Rule Them All: \\Generating LO-Shot Soft-Label Prototypes}


\author{\IEEEauthorblockN{Ilia Sucholutsky\IEEEauthorrefmark{1},
Nam-Hwui Kim\IEEEauthorrefmark{2}, Ryan P. Browne\IEEEauthorrefmark{3} and
Matthias Schonlau\IEEEauthorrefmark{4}}
\IEEEauthorblockA{Department of Statistics and Actuarial Science,
University of Waterloo,\\
Waterloo, Canada\\
Email: \IEEEauthorrefmark{1}isucholu@uwaterloo.ca,
\IEEEauthorrefmark{2}namhwui.kim@uwaterloo.ca,
\IEEEauthorrefmark{3}ryan.browne@uwaterloo.ca,
\IEEEauthorrefmark{4}schonlau@uwaterloo.ca}}




\maketitle

\begin{abstract}
	Increasingly large datasets are rapidly driving up the computational costs of machine learning. Prototype generation methods aim to create a small set of synthetic observations that accurately represent a training dataset but greatly reduce the computational cost of learning from it. Assigning soft labels to prototypes can allow increasingly small sets of prototypes to accurately represent the original training dataset. Although foundational work on `less than one'-shot learning has proven the theoretical plausibility of learning with fewer than one observation per class, developing practical algorithms for generating such prototypes remains an unexplored territory. We propose a novel, modular method for generating soft-label prototypical lines that still maintains representational accuracy even when there are fewer prototypes than the number of classes in the data. In addition, we propose the Hierarchical Soft-Label Prototype k-Nearest Neighbor classification algorithm based on these prototypical lines. We show that our method maintains high classification accuracy while greatly reducing the number of prototypes required to represent a dataset, even when working with severely imbalanced and difficult data. Our code is available at \url{https://github.com/ilia10000/SLkNN}.
\end{abstract}


\section{Introduction}
`Less than one'-shot (LO-shot) learning is a recently proposed setting wherein a model must learn to recognize $N$ classes from only $M<N$ training examples\cite{sucholutsky2020less}. The underlying premise is that this extreme level of data-efficiency may be attainable by assigning richer labels (or annotations) to training examples. In particular, it was analytically proven that it is possible to generate a small number $M$ of soft-label prototypes such that a kNN classifier fitted on these prototypes could discern $N>M$ classes. Reducing the number of prototypes required to represent a training dataset is especially valuable for instance-based algorithms like kNN as the computational complexity at inference time depends on the number of training examples the classifier was fitted on. Unexpectedly, \cite{sucholutsky2020less} showed in their main theorem that with just two carefully-designed soft-label prototypes it is possible to separate any finite number of classes. However, this particular result requires that the classes being separated lie roughly on a 1-dimensional manifold so that a soft-label prototype could be assigned to each end of the manifold and used to separate the classes in between. 

While the results above have shown that LO-shot learning is theoretically plausible, a practical algorithm for harnessing its potential is yet to be developed. Thus, we propose the first method for generating prototypical lines and a new classification algorithm that can use them to perform LO-shot learning in practice. Our key contributions can be summarized as follows:
\begin{itemize}
    \item We develop three methods for finding co-linear classes.
    \item We develop a method for producing `prototypical lines' by optimizing the two soft-label prototypes assigned to each set of approximately co-linear classes.
    \item We develop a novel classification algorithm, the Hierarchical Soft-Label Prototype kNN (HSLaPkNN), that can use the prototypical lines produced by these two methods. 
    \item We show that HSLaPkNN can perform LO-shot learning with prototypical lines and determine the tradeoff between dataset size reduction and classification accuracy. In particular, our method can retain over 90\% of the classification accuracy of 1NN while reducing the required number of prototypes (prototypical lines) by up to 80\%.
\end{itemize}
Our modular approach allows newly-developed algorithms for each component to be swapped in without interfering with the remaining components. This implies that our approach accommodates a continuous performance improvement through component-wise innovations. This is indeed an attractive feature for researchers, as the progress on individual components can by combined into the improvement of overall process.
The remainder of this paper is divided into four sections. In Section~\ref{algo} we detail our method, each of its components, and the theory behind them. In Section~\ref{exps} we describe our experimental setup and results. In Section~\ref{related} we discuss previous work in related areas. In Section~\ref{con} we analyze the impact of our method and suggest promising directions for future work.

\section{LO-Shot Prototype Generation Algorithm}
\label{algo}
\begin{figure}
    \centering
    \begin{subfigure}{0.45\textwidth}
    \caption{Select a dataset}
    \vspace{-3mm}
    \includegraphics[width=\textwidth]{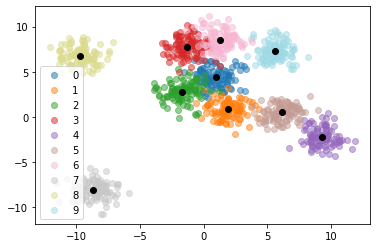}
    \end{subfigure}
    \begin{subfigure}{0.45\textwidth}
    \caption{Find lines corresponding to 1D manifolds}
    \vspace{-3mm}
    \includegraphics[width=\textwidth]{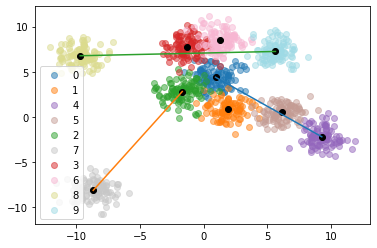}
    \end{subfigure}
    \begin{subfigure}{0.45\textwidth}
    \centering
    \caption{Solve system to find two soft-label prototypes for each line}
    \vspace{-2.5mm}
    \includegraphics[width=0.4\textwidth]{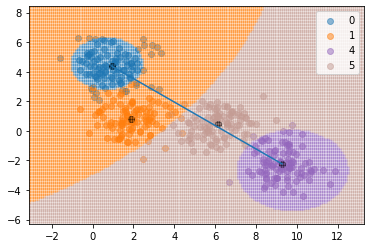}
    \includegraphics[width=0.415\textwidth]{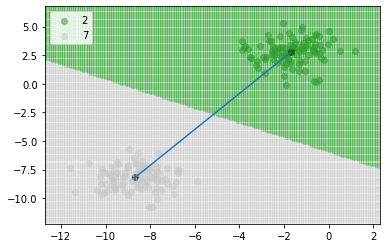}
    \includegraphics[width=0.4\textwidth]{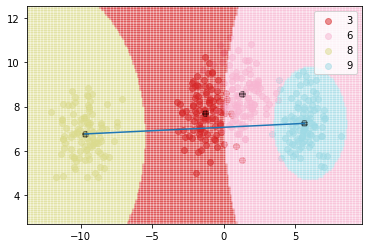}
    \end{subfigure}
    \begin{subfigure}{0.45\textwidth}
    \caption{Classify points based on nearest line and its two prototypes}
    \vspace{-3mm}
    \includegraphics[width=\textwidth]{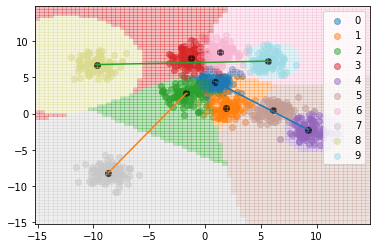}
    \end{subfigure}
    \caption{Prototype generation and classification process Hierarchical Soft-Label Prototype k-Nearest Neighbors (HSLaPkNN)}
    \label{fig:ourprocess}
\end{figure}

Our method primarily consists of three modular components. In the first component, we find the prototypical lines where each line contains information on a subset of original classes and every class belongs to one of the lines. Then, the second component finds optimal prototypes for every class associated with each line. Finally, the HSLaPkNN algorithm uses the lines and prototypes to classify the dataset. In particular, we present three different algorithms for finding the prototypical lines. We present a visual illustration of each component in Figure~\ref{fig:ourprocess}.

\subsection{Component 1: Finding Lines}
The objective of the first component of our prototype generation method is to find subsets of classes that lie along the same 1-dimensional manifold. We can find subsets of this type by grouping together classes that are approximately co-linear. In other words, our objective is roughly to find the smallest possible set of lines that cover (pass through) all classes.  We propose two methods for finding such lines.
\subsubsection{Brute Force}
One straightforward approach for finding the best combination of $M$ lines that pass through/near all the classes is to generate all such combinations of lines, score each one, and finally pick the best one. To reduce interactions between different lines in Component 3, we first filter out all combinations of lines where any of the lines intersect. For our experiments with this approach, we score each line by first finding all the classes that are closest to that line, and then taking the sum of the absolute values of the shortest distances from every point in those classes to the line. The scores are summed across each combination of lines, and the combination with the lowest score is chosen. This process is detailed in Algorithm~\ref{algo:brute}. Unfortunately, the complexity of this algorithm is $O(n^{2*l})$ where $n$ is the number of classes and $l$ is the number of lines. While computationally expensive, it is guaranteed to find near-optimal lines. As a result, this method is best used when there are either relatively few classes or only a small number of lines is required to cover all of them. For datasets with a large number of classes that cannot be covered with a small number of lines, we instead propose an approximate method.
\begin{algorithm}
\SetAlgoLined
\KwResult{Best set of M non-intersecting lines (as pairs of endpoints) for covering all classes}
 M = desired number of lines\;
 \textbf{C} = centroids of each class\;
 \textbf{Lines} = all combinations of two elements of \textbf{C}\; 
 \textbf{M\_Lines} = all combinations of M elements of \textbf{Lines}\; 
 best\_lines = None\;
 min\_dist = -1\;
 \For{cur\_lines \textbf{in} \textbf{M\_Lines}}{
    \If{no\_intersections(cur\_lines)}{
        cur\_dist=0\;    
        \For{c \textbf{in} \textbf{C}}{
            nearest = nearest\_line(c, cur\_lines)\;
            cur\_dist += dist\_to\_line(c, nearest)\;
        }
        \If{cur\_dist < min\_dist \textbf{or} min\_dist==-1}{
            min\_dist = cur\_dist\;
            best\_lines = cur\_lines\;
        }
    }
 }
 \Return best\_lines\;
 \caption{Brute-force line-finding algorithm}
 \label{algo:brute}
\end{algorithm}
\subsubsection{Recursive Regression}

The exhaustive enumeration like the brute force approach can lose its viability quickly. Instead, a preliminary clustering of the classes can produce a more elegant algorithm for line-finding. Let $N$ be the number of classes present, and denote by $c_1, \ldots, c_N$ the class-wise centroids. We want to find $M$ lines. The set $\{c_1, \ldots, c_N\}$ is partitioned into $M$ clusters such that each cluster contains at least two centroids. Our paper uses hierarchical clustering with the single linkage \cite{everitt2011cluster}, but a different method could be deployed based on the investigator's judgement. For the purpose of regression fit, we selected the last feature of the dataset to be the `response' and the rest of the features to be the `covariate'. This means that, for a centroid $c$, its last entry could function as a response variate and the remaining entries would be the covariates. However, the response-covariate configuration may be different based on the context or other external information. As an illustration, consider one of the clusters $G_i$ with size $n_i$, $G_i = \{c_{i1}, c_{i2}, \ldots, c_{i n_i}\}$, where each element in $G_i$ is a centroid. We partition $G_i$ into $A_i = \{c_{i1}, c_{i2}\}$ and $B_i = G_i \setminus A_i$ where $c_{i1}$ and $c_{i2}$ are of maximum pairwise Euclidean norm between the centroids in $G_i$. We fit a regression line $\beta_i$ on $c_{i1}$ and $c_{i2}$. Then, for each $c \in B_i$, we estimate a line $\hat{\beta}$ on $A_i \cup \{c\}$. Then, if $||\beta - \hat{\beta}||_2$ is less than a pre-determined tolerance $\epsilon > 0$, then $c$ is added to $A_i$ and $B_i$ is updated to $B_i \setminus \{c\}$. Otherwise, $c$ is not added to $A_i$ and is discarded from $B_i$. This forward selection process is repeated until $B_i = \emptyset$. This procedure is applied to all $G_i$ for $i = 1, 2, \ldots, M$. The results are $\{A_1, \ldots, A_M\}$ and $\{\beta_1, \ldots, \beta_M\}$, where each line $\beta_j$ ($j = 1, \ldots, M$) is segmented to have endpoints at the furthest-apart pair of centroids that generated it. This method will be called Recursive Regression (RR) hereafter. 

\begin{algorithm}
\SetAlgoLined
\KwResult{A set of M lines and classes assigned to each line}
 \textbf{C} = centroids of each class\;
 M = number of preliminary clusters that contain at least two different centroids from \textbf{C}\;
 $\textbf{G}_i$ (i = 1, 2, \ldots, M) = $i^\text{th}$ preliminary cluster\; 
 $\epsilon$ = pre-determined positive maximum tolerance\;
 best\_lines = $\emptyset$\;
 captured\_groups = $\emptyset$\;
 \For{i \textbf{in} 1, 2, \ldots, M}{
    \textbf{A} = set of two furthest centroids in $\textbf{G}_i$ in terms of Euclidean distance\;
    \textbf{B} = $\textbf{G}_i \setminus \textbf{A}$\;
    $\beta$ = regression line fitted on two furthest centroids in $\textbf{G}_i$\;
    \While{\textbf{B} $\neq \emptyset$} {
    all\_dist = $\{||\beta - \beta_{\textbf{A} \cup \{c\}}||_2\}_{c \in \textbf{B}}$ where $\beta_{\textbf{A} \cup \{c\}}$ is the regression line fitted on $\textbf{A} \cup \{c\}$\;
        \If{all(all\_dist > $\epsilon$)} {
        \textbf{stop}
        }
        $c^* = $ minimizer among $\textbf{B}$ of all\_dist \;
        \textbf{A} = \textbf{A} $\cup$ $c^*$\;
        \textbf{B} = \textbf{B} $\setminus$ $c^*$\;
    }
    best\_lines = best\_lines $\cup$ $\beta$\;
    captured\_groups = captured\_groups $\cup$ $\{$classes\_associated\_with\_\textbf{A}$\}$\;
 }
 \Return best\_lines, captured\_groups\;
 \caption{Recursive Regression (RR) for line-finding}
 \label{algo:recreg}
\end{algorithm}

\subsubsection{Distance-based Attraction}
While the Recursive Regression is careful when adding classes to the distilling lines, a possible shortfall is its dependence on the initial clustering result, because the cluster-wise regression lines are not altered. Moreover, the tolerance threshold $\epsilon$ can influence the number of classes that are left out. To mitigate these issues, we propose another clustering-based algorithm called Distance-based Attraction (DA). The initial clustering stage is the similar to that of RR, and we obtain $M$ many line segments $\beta_1, \ldots, \beta_M$. Then, for each centroid $c_i$ ($i = 1,\ldots, N$), we compute the shortest distance from it to each line segment, denoted by $d_{\beta_1}, \ldots, d_{\beta_M}$. Then, $c_i$ is assigned to the line $\text{argmin}_{\beta_1, \ldots, \beta_M}\{ d_{\beta_1}, \ldots, d_{\beta_M} \}$. Notice that we do not require a tolerance threshold, and that every class is guaranteed to be assigned to a line. Once the assignment is completed, the line segment for each cluster is re-calculated, as there may be a new furthest-apart pair of centroids.

\begin{algorithm}
\SetAlgoLined
\KwResult{A set of M lines and classes assigned to each line}
 \textbf{C} = $\{c_1, \ldots, c_N \}$ = centroids of each class\;
 M = number of preliminary clusters that contain at least two different centroids from \textbf{C}\;
 $\textbf{G}_i$ (i = 1, 2, \ldots, M) = $i^\text{th}$ preliminary cluster\; 
 pre\_lines = line segments generated from furthest-apart pair of centroids from each $\textbf{G}_i$\;
 best\_lines = $\emptyset$\;
 group\_assignment = $\emptyset$\;
 \For{i \textbf{in} 1, 2, \ldots, N}{
    index = $\underset{j = 1,\ldots, M}{\text{argmin}}$($\text{shortest\_dist}(c_i, \beta_j)$)\;
    group\_assignment = group\_assignment $\cup$ $\{\text{index}\}$\;
 }
 \For {i in unique\_elements(group\_assignment)} {
    set = $\{c_j : \text{group\_assignment}[j-1] = i \}$\;
    $\beta_i^*$ = line\_on\_furthest\_pair(set)\;
    best\_lines = best\_lines $\cup$ $\beta_i^*$\;
 }
 \Return best\_lines, group\_assignment\;
 \caption{Distance-based Attraction (DA) for line-finding}
 \label{algo:DA}
\end{algorithm}

\subsection{Component 2: Finding Optimal Prototypes for a Single Line}
Once a suitable line segment is found, along with the classes assigned to it, we can use the idea of the main theorem from~\cite{sucholutsky2020less} to design two soft-label prototypes that will be placed at each endpoint of the line segment. The soft labels of these two prototypes must be designed in such a way that a SLaPkNN classifier fitted on them would accurately separate the classes lying along the line segment. Similarly to~\cite{sucholutsky2020less} we can formulate this as an optimization problem where we want to maximize each class's influence over its interval of the line segment. We approximate each class's interval of the line segment as starting from the midpoint between the class centroid and the preceding class's centroid, and ending at the midpoint between the class centroid and the next class's centroid. We approximate a class's influence over its interval by its influence at the centroid of that interval. A class's influence at a given point is equal to the sum of the associated soft-label value divided by distance from the prototype, for each prototype. For a point to be assigned to a particular class, that class's influence must be higher than all the other classes' influences at that point. To enforce this, at each interval centroid we not only add a constraint forcing the desired class to have the highest influence, but we also actually maximize the \textit{difference} between the influence of the desired classes and the sum of the influences of all the other classes. A key difference between the system we aim to solve and the one solved in~\cite{sucholutsky2020less} is that we do not assume that classes will be distributed symmetrically along the line segment. As a result, we cannot use the simplifying constraints that the soft labels of the two prototypes are symmetrical. Instead, we add the additional constraint that the influence of neghboring classes must be equal at the midpoint of their centroids. In order to solve the resulting optimization problem we use the CVXPY library~\cite{diamond2016cvxpy, agrawal2018rewriting}. The full algorithm for generating and solving this optimization problem is specified in Algorithm~\ref{algo:syseq}.

\begin{algorithm}
\SetAlgoLined
\KwResult{Two lists containing the soft labels corresponding to the two prototypes}
 p1 = location of first prototype\;
 p2 = location of second prototype\;
 lineseg = [p1, p2]\;
 centroids = centroids of each class assigned to line\;
 N = number of centroids assigned to line\;
 x = length 2N array of variables to optimize\;
 $\epsilon$ = 0.01\;
 projections = []\;
 dists = []\;
 middists=[]\;
 \For{centroid \textbf{in} centroids}{
    projection = proj(centroids, lineseg)\;
    projections.append(projection)\;
    dist = dist(p1, projection)\;
    dists.append(dist)\;
    middists.append(dist/2)\;
 }
 A=[]\;
 constraints=[]\;
 \For{i \textbf{in} 0, 1, 2, \ldots, N-1}{
    vector = zeros(2*N) \;
    vector[i] += 1/(dists[i]+$\epsilon$-p1) \;
    vector[N+i] += 1/(p2-dists[i]+$\epsilon$) \;
    q1 = x[i]/(dists[i]+$\epsilon$-p1)\;
    q2 = x[N+i]/(p2-dists[i]+$\epsilon$)\;
    \For{j \textbf{in} 0, 1, 2, \ldots, N-1}{
        \If{i $\neq$ j}{
            vector[j] -= 1/(dists[i]+$\epsilon$-p1)\;
            vector[N+j] -= 1/(p2-dists[i]+$\epsilon$)\;
            q3=x[j]/(dists[i]+$\epsilon$-p1)\;
            q4=x[N+j]/(p2-dists[i]+$\epsilon$)\;
            constraint = q1+q2>=q3+q4+$\epsilon^2$\;
            constraints.append(constraint)\;
        }
    A.append(vector)\;
    }
    \If{i<N-1}{
        q1 = x[i]/(mid\_dists[i+1]-p1)\;
        q2 = x[N+i]/(p2-mid\_dists[i+1])\;
        q3=x[i+1]/(mid\_dists[i+1]-p1)\;
        q4=x[N+i+1]/(p2-mid\_dists[i+1])\;
        constraint = q1+q2==q3+q4\;
        constraints.append(constraint)\;
    }
 }
constraints.append(x>=0)\;
constraints.append(x<=1)\;
constraints.append(sum(x[0:N])==1)\;
constraints.append(sum(x[N:2N])==1)\;
objective = Maximize(sum(A'x)+sum\_smallest(A'x,2))\;
result = solve(objective, constraints)\;

 \Return x.value[0:N], x.value[N:2N];
 \caption{Generating system of equations and constraints for two soft-label prototypes}
 \label{algo:syseq}
\end{algorithm}

\subsection{Component 3: Classifying with Multiple Lines}
The two prototypes assigned to the endpoints of a single line are near-optimal for fitting a SLaP2NN classifier if that line and its associated classes are isolated from all other classes. However, in practice, lines could pass fairly close to each other. As a result, the two nearest prototypes to a particular point on a line may not end up being the two prototypes assigned to the endpoints of that line. In order to rectify this problem, we propose the Hierarchical Soft-Label Prototype kNN (HSLaPkNN) classification rule. This classifier performs two main steps when determining how to classify a target point. First, it finds the nearest prototype line to the target point. Second, it fits a SLaP2NN classifier on the two endpoint prototypes assigned to that line. These steps are detailed in Algorithm~\ref{algo:HSLaPkNN}. We note that the algorithm is intentionally designed to allow more than the minimum required two prototypes per line in case the user wishes to improve accuracy by adding additional prototypes.

\begin{algorithm}
\SetAlgoLined
\KwResult{Predicted classes of every target point in $P$}
 $P$ = list of target points for classification\;
 lines = list of M pairs of prototypes\;
 Preds = []\;
 \For{$p$ \textbf{in} $P$}{
     nearest = nearest\_line($p$, lines)\;
     soft\_pred = [0]*N\;
     \For{prototype \textbf{in} nearest}{
         proto\_loc = prototype[0]\;
         proto\_lab = prototype[1]\;
         dist = dist(proto\_loc, $p$)\;
         soft\_pred += proto\_lab/dist\;
     }
     hard\_pred = argmax(soft\_pred)\;
     Preds.append(hard\_pred)\;
 }
 \Return Preds\;
 \caption{Hierarchical Soft-Label Prototype k-Nearest Neighbor (HSLaPkNN) classification rule}
 \label{algo:HSLaPkNN}
\end{algorithm}

\section{Experiments}
\label{exps}
\begin{figure}[htbp!]
    \centering
    \begin{subfigure}{0.45\textwidth}
    \caption{Example result of Regular2 experiment}
    \vspace{-3mm}
    \includegraphics[width=\textwidth]{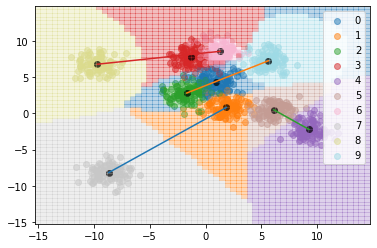}
    \end{subfigure}
    \begin{subfigure}{0.45\textwidth}
    \caption{Example result of Regular (5) experiment}
    \vspace{-3mm}
    \includegraphics[width=\textwidth]{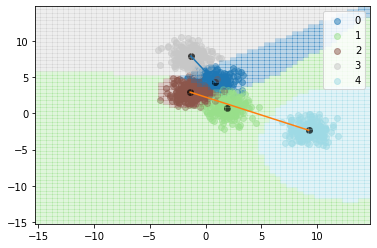}
    \end{subfigure}
    \begin{subfigure}{0.45\textwidth}
    \caption{Example result of Imbalanced1 experiment}
    \vspace{-3mm}
    \includegraphics[width=\textwidth]{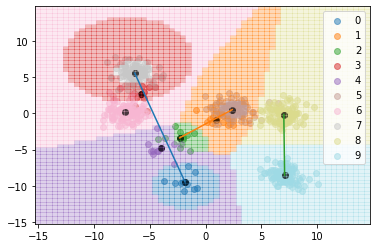}
    \end{subfigure}
    \begin{subfigure}{0.45\textwidth}
    \caption{Example result of Imbalanced2 experiment}
    \vspace{-3mm}
    \includegraphics[width=\textwidth]{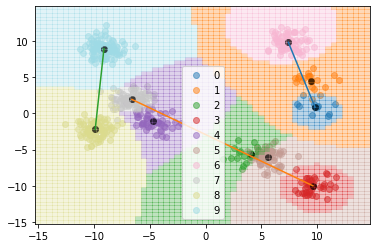}
    \end{subfigure}
    \caption{Examples of resulting HSLaPkNN decision landscapes.}
    \label{fig:results1}
\end{figure}

\begin{figure}[htbp!]
    \centering
    \begin{subfigure}{0.45\textwidth}
    \caption{Example result of Giant experiment with 25 lines found}
    \vspace{-3mm}
    \includegraphics[width=\textwidth]{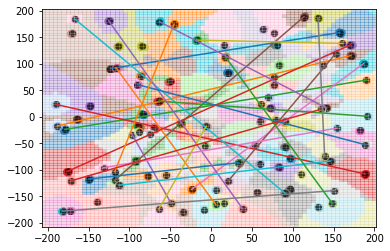}
    \end{subfigure}
    \begin{subfigure}{0.45\textwidth}
    \caption{Example result of Giant experiment with 30 lines found}
    \vspace{-3mm}
    \includegraphics[width=\textwidth]{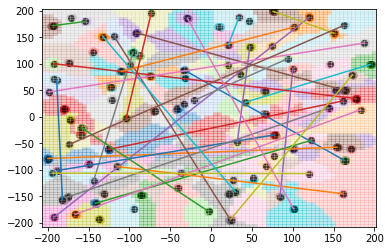}
    \end{subfigure}
    \caption{Examples of resulting HSLaPkNN decision landscapes.}
    \label{fig:results2}
\end{figure}

We perform a variety of experiments to determine the tradeoff between classification accuracy and dataset size reduction offered by our prototyping algorithm and HSLaPkNN. For every experiment, we also fit a normal 1NN classifier as a baseline. Each type of experiment is summarized below and some examples of the resulting classification decision boundaries are visualized in Figures~\ref{fig:results1} and~\ref{fig:results2}.
\begin{itemize}
    \item Regular1: Brute force line-finding is used to find three lines in a 10-class dataset consisting of 1000 points with two feature dimensions. Each class consists of 100 points.
    \item Regular2: Brute force line-finding is used to find four lines in a 10-class dataset consisting of 1000 points with two feature dimensions. Each class consists of 100 points.
    \item Regular (5): Brute force line-finding is used to find two lines in a 5-class dataset consisting of 1000 points with two feature dimensions. Each class consists of 200 points.
    \item Giant: Distance-based attraction is used to find some number of lines that cover all classes in a 100-class dataset consisting of 2000 points with two feature dimensions. Each class consists of 20 points. For this experiment we also record the average number of lines found along with the other metrics.
    \item Imbalanced1: Brute force line-finding is used to find three lines in a 10-class dataset consisting of 550 points with two feature dimensions. Five classes consist of 10 points each, and five classes consist of 100 points each. 
    \item Imbalanced2: Brute force line-finding is used to find three lines in a 10-class dataset consisting of 550 points with two feature dimensions. Class $i$ consists of $10i$ points for $i=1,2,...,10$.
    \item Small: Brute force line-finding is used to find three lines in a 10-class dataset consisting of 100 points with two feature dimensions. Each class consists of 10 points.
    \item Penguins: Distance-based attraction is used to find some number of lines that cover all classes in the 5-class version of the Palmer Penguins dataset~\cite{penguins}. We use the four continuous explanatory variables (bill length,  bill depth,   flipper length, body mass) as the features, and the combination of `species' and `island' as the class. 
    \item EColi: Distance-based attraction is used to find some number of lines that cover all classes in the 5-class version of the E. Coli dataset (`ecoli') from OpenML~\cite{OpenML2013}. We use the six continuous explanatory variables (mcg, gvh, lip, aac, alm1, alm2) as the features.
\end{itemize}

We repeat each experiment involving simulated data 100 times with a different random seed and record the mean and standard deviation of the classification accuracy in Table~\ref{tab:Results1} along with other details about each experiment.  In order to understand the tradeoff between dataset size reduction and classification accuracy, we also calculate the ratio of the number of prototypical lines used by HSLaPkNN compared to 1NN, as well as the ratio of their classification accuracies when fitted on these prototypes. We summarize these results in Table~\ref{tab:Ratios1}. Notably, our method retains upwards of 90\% of the classification accuracy of 1NN while reducing the number of nearest prototypes (prototypical lines) that must be considered at inference time by as much as 80\%.

Our method is most interpretable when working with datasets that have two-dimensional feature sets due to the ease of visualizing the resulting decision landscapes. However, our method can also work with higher-dimensional datasets as seen by the results with the Palmer Penguins dataset and E. Coli dataset. In order to better understand the effect of data dimensionality on the performance of our method, we perform an additional set of experiments where we hold all other hyperparameters of the algorithm and data-generation constant but increase the dimensionality of the datasets. Table~\ref{tab:Results1} summarizes the mean and standard deviations of the classification accuracy achieved by HSLaPkNN and vanilla 1NN on these experiments with synthetic datasets containing 80 classes. Table~\ref{tab:Ratios2} summarizes the associated prototype and accuracy ratios. Each experiment uses the distance-based attraction line-finding method and is repeated 100 times with a different random seed used to generate the data each time. Because we hold all hyperparameters constant, our method exhibits lower classification accuracy on this set of experiments. Curiously, we notice that our method exhibits greater variability in classification accuracy when used with higher-dimensional datasets. We believe that this may be caused by the sparsity introduced at higher dimensions (i.e. the curse of dimensionality) but further investigation is required to confirm this.

\begin{table}[]
    \centering
    \begin{tabular}{|c|c|c|c|c|}
    \hline
    Experiment & Points & Lines & HSLaPkNN ($\mu \pm \sigma$)& 1NN ($\mu \pm \sigma$)\\
    \hline
       Regular &1000 & 3 & 0.814 $\pm$ 0.063 & 0.895 $\pm$ 0.045 \\
       Regular &1000 & 4 & 0.837 $\pm$ 0.06 & 0.894 $\pm$ 0.045 \\
       Regular (5) & 1000 & 2 & 0.909 $\pm$ 0.077 & 0.951 $\pm$ 0.05\\
       Small & 100 & 3 & 0.82 $\pm$ 0.06 & 0.90 $\pm$ 0.05 \\
       Imbalanced1 & 550 & 3 & 0.815 $\pm$ 0.098 & 0.896 $\pm$ 0.059\\
       Imbalanced2 & 550 & 3 & 0.813 $\pm$ 0.081 & 0.895 $\pm$ 0.054\\
       Giant & 2000 & 23.82 & 0.835 $\pm$ 0.042 & 0.996 $\pm$ 0.004\\
       Penguins & 342 & 1 & 0.4327 & 0.5029 \\
       EColi & 327 & 2 & 0.8135 & 0.8654 \\
       \hline
    \end{tabular}
    \caption{Experimental results on a variety of simulated datasets comparing the performance of HSLaPkNN fitted on our soft-label prototypes to vanilla 1NN fitted on class centroids. Lines refers to the average number of lines found for the dataset. Experiments involving synthetic data (all except Penguins and EColi) are repeated 100 times with different random seeds during data generation to produce the standard deviations.}
    \label{tab:Results1}
\end{table}

\begin{table}[]
    \centering
    \begin{tabular}{|c|c|c|c|c|}
    \hline
    Experiment & Points & Lines & Prototypes ratio & Accuracy ratio\\
    \hline
       Regular &1000 & 3 & 0.3 & 0.91 \\
       Regular &1000 & 4 & 0.4 & 0.936 \\
       Regular (5) & 1000 & 2 & 0.2 & 0.948\\
       Small & 100 & 3 & 0.3 & 0.911 \\
       Imbalanced1 & 550 & 3 & 0.3 & 0.909\\
       Imbalanced2 & 550 & 3 & 0.3 & 0.909\\
       Giant (100) & 2000 & 23.82 & 0.238 & 0.838\\
       Penguins & 342 & 1 & 0.2 & 0.86\\
       EColi & 327 & 2 & 0.4 & 0.94\\
       \hline
    \end{tabular}
    \caption{Experimental results on a variety of simulated datasets comparing the performance of HSLaPkNN fitted on our soft-label prototypes to vanilla 1NN fitted on class centroids. Prototypes ratio refers to the ratio of the number of prototypical lines used by HSLaPkNN to the number of prototypes used by 1NN. Accuracy ratio refers to the ratio of the mean classification accuracy of HSLaPkNN to the mean classification accuracy of 1NN.}
    \label{tab:Ratios1}
\end{table}

\begin{table}[]
    \centering
    \begin{tabular}{|c|c|c|c|}
    \hline
        Dimension & Lines & HSLaPkNN ($\mu \pm \sigma$) & 1NN ($\mu \pm \sigma$) \\
    \hline
        2 & 23.3 & 0.497 $\pm$ 0.034 & 0.782 $\pm$ 0.021\\
        3 & 22.7 & 0.692 $\pm$ 0.046 & 0.974 $\pm$ 0.01\\
        4 & 22.6 & 0.717 $\pm$ 0.055 & 0.997 $\pm$ 0.002\\
        5 & 22.0 & 0.701 $\pm$ 0.062 & 0.999 $\pm$ 0.001 \\
        6 & 22.3 & 0.692 $\pm$ 0.068 & 1 $\pm$ 0\\
        7 & 21.8 & 0.67 $\pm$ 0.067 & 1 $\pm$ 0.0 \\
        8 & 21.4 & 0.653 $\pm$ 0.062 & 1 $\pm$ 0.0 \\
        9 & 21.5 & 0.637 $\pm$ 0.073 & 1 $\pm$ 0.0 \\
        10 & 21.5 & 0.63 $\pm$ 0.07 & 1 $\pm$ 0.0 \\
        \hline
    \end{tabular}
    \caption{Experimental results on simulated datasets of different dimensionalities comparing the performance of HSLaPkNN fitted on our soft-label prototypes to vanilla 1NN fitted on class centroids. Each dataset contains 2000 points across 80 classes. Experiments are repeated 100 times with different random seeds during data generation to produce the standard deviations.}
    \label{tab:Results2}
\end{table}

\begin{table}[]
    \centering
    \begin{tabular}{|c|c|c|c|}
    \hline
        Dimension & Lines & Prototypes ratio & Accuracy ratio\\
    \hline
        2 & 23.3 & 0.291 & 0.635\\
        3 & 22.7 & 0.284 & 0.71\\
        4 & 22.6 & 0.283 & 0.719\\
        5 & 22.0 & 0.275 & 0.701 \\
        6 & 22.3 & 0.279 & 0.692 \\
        7 & 21.8 & 0.273 & 0.67 \\
        8 & 21.4 & 0.268 & 0.653 \\
        9 & 21.5 & 0.269 & 0.637\\
        10 & 21.5 & 0.269 & 0.63 \\
        \hline
    \end{tabular}
    \caption{Experimental results on simulated datasets with increasing feature dimensionalities comparing the performance of HSLaPkNN fitted on our soft-label prototypes to vanilla 1NN fitted on class centroids. Each dataset contains 2000 points across 80 classes. Prototypes ratio refers to the ratio of the number of prototypical lines used by HSLaPkNN to the number of prototypes used by 1NN. Accuracy ratio refers to the ratio of the mean classification accuracy of HSLaPkNN to the mean classification accuracy of 1NN.}
    \label{tab:Ratios2}
\end{table}

\section{Related Work}
\label{related}
\subsection{Finding co-linear classes}
The search for observations lying on a line can be dated back to the analysis of multicollinearity in linear regression \cite{belsley2005regression}. Conventionally, multicollinearity is a topic of concern in modelling due to it resulting in a verbose model. However, identifying co-linear observations could be useful in finding an efficient representation. If the classes in a dataset could be grouped by various co-linear structures, then a representation of arbitrarily many classes using a much smaller number of lines may be possible. This is the motivation behind the RR and DA algorithms presented in this paper. There has been significant past work on covering points with various geometrical objects \cite{langerman2001covering, langerman2005covering,grantson2006covering,gencc2011covering,dumitrescu2015approximability,carmi2007covering,mahapatra2007covering,ahn2011covering}. Our approach draws contrast from these methods in that we seek to use lines as a satisfactory approximation of a multi-class dataset, instead of a precise covering of all points.


\subsection{Prototype selection and generation}

Dataset Distillation (DD), the process of reducing a large dataset into a small sample of representative observations, has paved the way toward ``learning more from less''. In one of the most recent advances in DD, \cite{sucholutsky2019soft} showed that, with soft labelling, learning the classes in a dataset from fewer than one observation per class is possible. In particular, Soft-Label Dataset Distillation (SLDD) was used to create a dataset of just five distilled images, which is less than one per class, that trained neural networks to over 90\% accuracy on MNIST. While there exist a range of methods on selecting or generating prototypes from large data sets such as \cite{active1, active2, coreset3, coreset1, coreset2, bezdek2001nearest,triguero2011taxonomy, garcia2012prototype, kusner2014stochastic}, the novelty in our work lies in the development of practical algorithms for generating LO-shot prototypes using a small number of simple geometric objects (ie. straight line segments) to distill a large number of classes, thereby enabling the discovery of unseen classes in a straightforward manner. 

\section{Conclusion}
\label{con}
We have proposed an algorithm for finding LO-shot prototypes in practice. The algorithm is intentionally designed to be modular so that each component can be improved independently. Next steps include finding better algorithms for detecting co-linear classes in datasets, improving the formulation of the soft-label optimization problem, and generalizing the method to work with a greater variety of classifiers.

Our proposed algorithm currently makes distributional assumptions about the datasets and classes to which it is applied. In particular, it assumes that each class is fairly contiguous and disjoint. When these assumptions are violated, even existing hard-label prototype methods need to increase the number of prototypes they produce in order to maintain classification accuracy. \cite{sucholutsky2020optimal} dissect this phenomenon for a particularly pathological case where the number of hard-label prototypes required to represent a dataset may be quadratic in the number of classes and \cite{sucholutsky2020less} show that the required number of soft-label prototypes is constant. We believe an interesting direction would be to relax these assumptions for our soft-label prototype generation algorithm, perhaps by treating clusters in the data as sub-classes and optimizing for them separately rather than treating the entire class in a monolithic way. While this would likely increase the average number of classes assigned to each line, and may even increase the total number of lines required to achieve good coverage, it would likely result in higher performance on a larger variety of datasets.  

We note that our proposed algorithm builds directly on the result from \cite{sucholutsky2020less} regarding classes lying on a 1-dimensional manifold. However, it is possible that the underlying theory could be extended to classes lying on higher dimensional manifolds. In particular, we conjecture that, given some distributional assumptions, if a finite set of classes lies on an M-dimensional manifold, only M+1 soft-label prototypes are required to separate them. If this conjecture holds, then our proposed algorithm could be extended to work with M-dimensional manifolds. However, the key problem that would need to be solved is how to automatically detect subsets of the training dataset that lie on various manifolds with differing dimensionalities, and then optimize the selection of these subsets so as to minimize the total number of soft-label prototypes required to represent the dataset. When optimizing soft labels for these higher-dimensional manifolds, maintaining stability and robustness to noise may become increasingly important as there may be many more potentially unstable solutions than in the 1D case. Thus the optimization problem may require either a secondary objective that rewards stability or additional constraints that try to enforce it directly.

\bibliographystyle{unsrt}
\bibliography{references}  


\end{document}